\PassOptionsToPackage{unicode}{hyperref}
\PassOptionsToPackage{hyphens}{url}
\documentclass[
]{article}
\usepackage{amsmath,amssymb}
\usepackage{iftex}
\ifPDFTeX
  \usepackage[T1]{fontenc}
  \usepackage[utf8]{inputenc}
  \usepackage{textcomp} 
\else 
  \usepackage{unicode-math} 
  \defaultfontfeatures{Scale=MatchLowercase}
  \defaultfontfeatures[\rmfamily]{Ligatures=TeX,Scale=1}
\fi
\usepackage{lmodern}
\ifPDFTeX\else
\fi
\IfFileExists{upquote.sty}{\usepackage{upquote}}{}
\IfFileExists{microtype.sty}{
  \usepackage[]{microtype}
  \UseMicrotypeSet[protrusion]{basicmath} 
}{}
\makeatletter
\@ifundefined{KOMAClassName}{
  \IfFileExists{parskip.sty}{%
    \usepackage{parskip}
  }{
    \setlength{\parindent}{0pt}
    \setlength{\parskip}{6pt plus 2pt minus 1pt}}
}{
  \KOMAoptions{parskip=half}}
\makeatother
\usepackage{xcolor}
\usepackage[margin=3cm]{geometry}
\usepackage{longtable,booktabs,array}
\usepackage{calc} 
\usepackage{etoolbox}
\makeatletter
\patchcmd\longtable{\par}{\if@noskipsec\mbox{}\fi\par}{}{}
\makeatother
\IfFileExists{footnotehyper.sty}{\usepackage{footnotehyper}}{\usepackage{footnote}}
\makesavenoteenv{longtable}
\providecommand{\tightlist}{%
  \setlength{\itemsep}{0pt}\setlength{\parskip}{0pt}}
\usepackage{graphicx}
\usepackage{authblk}
\usepackage[numbers,square,sort&compress]{natbib}
\ifLuaTeX
  \usepackage{selnolig}  
\fi
\IfFileExists{bookmark.sty}{\usepackage{bookmark}}{\usepackage{hyperref}}
\IfFileExists{xurl.sty}{\usepackage{xurl}}{} 
\hypersetup{
  pdftitle={Evaluating Large Language Model Performance on International Maritime Dangerous Goods Code Compliance},
  pdfauthor={Alexander Thomas, Hubert P.H. Shum, Darren Nellis, Manli Zhu, Phatpicha Yochum, William Bartle, Daniel Wrightson},
  hidelinks,
  pdfcreator={LaTeX via pandoc}}

\title{Evaluating Large Language Model Performance on International Maritime Dangerous Goods Code Compliance}
\author[1]{Alexander Thomas\thanks{Corresponding author: alex.thomas@hazcheck.com}}
\author[2]{Hubert P. H. Shum\thanks{Corresponding author: hubert.shum@durham.ac.uk}}
\author[1]{Darren Nellis}
\author[1]{Manli Zhu}
\author[1]{Phatpicha Yochum}
\author[1]{William Bartle}
\author[1]{Daniel Wrightson}
\affil[1]{NCB Hazcheck Limited, UK}
\affil[2]{Durham University, UK}
\date{2026}

\begin{document}
\maketitle
\begin{abstract}
The transport of dangerous goods by sea is a high-consequence activity governed by the International Maritime Dangerous Goods (IMDG) Code\footnote{The IMDG Code is copyrighted by the IMO. References to the Code throughout this paper provide background and context for the study.}, a complex regulatory framework where errors in classification, packaging, stowage, or segregation can result in fire, explosion, toxic release, or loss of life or vessel. Correct compliance requires accurately interpreting hundreds of pages of interacting provisions, updated on a two-year amendment cycle. Practitioners increasingly use Large Language Models (LLMs) as decision-support tools, yet no systematic evaluation exists of whether they can reliably interpret IMDG requirements for safety-critical use.

This paper introduces DGEval, the first benchmark for evaluating LLM knowledge of IMDG Amendment 42-24. Built from expert-written questions on a commercial e-learning platform and structured lookups from the Dangerous Goods List (DGL), it comprises 1,678 questions across multiple-choice, open-ended, DGL lookup, and regulatory identification tasks. We evaluate 13 models from six providers across multiple thinking configurations, including one maritime domain-specific fine-tuned model, and test the effect of web search.

Although the best-performing model exceeds the human practitioner baseline on multiple-choice questions, all models are weakest in the operationally safety-critical areas of stowage, segregation, and regulatory recall. These results indicate that LLMs may support compliance tasks, particularly structured DGL lookups with web search, but unreliability in operational areas and regulatory-text recall means human oversight and authoritative source verification remain necessary before deployment in any safety-critical context.
DGEval is designed as a safety assurance instrument to be applied continuously as models evolve, not as a settled characterisation of current capability.
\end{abstract}

\hypertarget{introduction}{%
\section{1. Introduction}\label{introduction}}

Maritime transport of dangerous goods operates at substantial scale: global container port
throughput reached approximately 800 million twenty-foot equivalent units (TEUs) in 2022
\citep{WorldBank2022}, of which an estimated 8\%\footnote{Hazcheck internal estimate}
contained some form of dangerous goods, a proportion expected to grow as
shipments of products containing lithium batteries increase.
The international carriage of these goods by sea is governed by the International Maritime Dangerous Goods (IMDG) Code,
which is mandatory for states party to the International Convention for the Safety of Life at Sea (SOLAS)
and applies across the great majority of global shipping.
Non-compliance can result in consequences that range from financial penalties to fire,
explosion, toxic release, or loss of life, vessel or cargo.
Once at sea, such incidents are nearly impossible to remediate
\citep{HumanErrorAtSea, ShippingIncidents2025}.

Establishing compliance for a single consignment is not a simple lookup task.
The IMDG Code spans hundreds of pages, and the Dangerous Goods List (DGL)
contains over 3,000 entries, each associated with regulatory properties
covering classification, documentation, packaging, marking, labelling, placarding, stowage, and segregation.
These properties are not independent: correct determination of stowage and segregation
requirements, the most operationally safety-critical decisions, requires multi-step
reasoning across several interacting DGL fields and Code sections.
The Code is revised every two years, so practitioners must also track amendment changes
to ensure they are applying the current version.
Human error is estimated to contribute to 96\% of maritime accidents \citep{HumanErrorAtSea},
making the accuracy of any compliance tool a direct safety concern.

Commercially available Large Language Models (LLMs) now demonstrate expert-level performance
in many professional domains \citep{wang2024mmluprorobustchallengingmultitask, rein2023gpqagraduatelevelgoogleproofqa}.
Because they answer open-ended natural-language queries, they are an appealing
alternative to manual consultation of the Code.
Their use has become widespread across professional and knowledge work \citep{liang2025widespread},
including regulated compliance checking \citep{li2025regtech},
and has reached the maritime and logistics sector: commissioned industry research documents a rapid
rise in AI adoption among shipping companies \citep{lr_thetius_2024}, and LLM
applications are emerging across supply chain management, including risk and compliance functions
\citep{song2026llmscm}. Within dangerous goods compliance specifically, NCB Hazcheck has received a growing number of
requests to verify AI-generated answers to DG queries,\footnote{NCB Hazcheck internal observation.}
and AI-generated DG guidance containing factual errors is already circulating in professional
forums \citep{LinkedInDG1, LinkedInDG2}.

Such mistakes are not harmless: an incorrect LLM answer, such as a wrong stowage code, missed segregation requirement,
or outdated packing instruction, can have catastrophic consequences at sea.
Strong average performance is insufficient as a deployment criterion in this context: a model that performs well overall may still fail systematically in the sub-domains where errors carry the most serious operational consequences.
Safe use therefore requires treating LLMs as decision-support tools, not decision-makers: no current model is fit to make DG compliance decisions unaided, so they must be deployed within governed workflows that retain trained practitioner oversight and re-evaluated continuously as models change.

Despite this, no systematic evaluation exists of LLM performance on DG compliance tasks.
General-purpose benchmarks such as MMLU-Pro \citep{wang2024mmluprorobustchallengingmultitask}
cover law, business, and science but make no reference to logistics, maritime regulation, or DG classification.
Domain-specific benchmarks have been developed for other knowledge-intensive domains, including
chemistry \citep{mirza2024largelanguagemodelssuperhuman}, finance \citep{islam2023financebenchnewbenchmarkfinancial},
and law \citep{guha2023legalbenchcollaborativelybuiltbenchmark}, but none addresses DG transport.
Nor is the gap filled on the model side: the only DG-adjacent system is LLaMarine \citep{nguyen2025llamarineopensourcemaritimeindustryspecific},
a fine-tuned model based on Llama 3.1 (70B) trained on broader maritime regulatory text;
it is not an evaluation benchmark and does not address DG compliance specifically.

This paper presents DGEval, the first benchmark designed specifically
to evaluate LLM knowledge of the IMDG Code under Amendment 42-24 (A42-24).
The benchmark is constructed from two proprietary sources: expert-written e-learning
content from the NCB Hazcheck platform and structured questions derived from
the DGL. It covers 1,678 questions across four complementary
sections. We evaluate 13 models (commercial, open-source, and one
domain-specific fine-tuned model) across multiple thinking
configurations, and test the impact of enabling web search.

Our contributions are:

\begin{enumerate}
\def\labelenumi{\arabic{enumi}.}
\tightlist
\item
  \textbf{DGEval}, the first DG-shipping benchmark grounded in IMDG
  A42-24, with expert-weighted scoring and human practitioner
  baselines, designed for repeated re-application as models change
  rather than as a one-time score.
\item
  \textbf{Comprehensive evaluation} of 13 models from six providers,
  with up to five thinking levels per model and a web-search
  sub-analysis.
\item
  \textbf{Actionable findings} for industry practitioners and benchmark
  designers, including model performance rankings, systematic weak areas, and
  cost-performance analysis.
\end{enumerate}

The remainder of this paper is structured as follows: \hyperref[background]{Section 2} provides
background on DG shipping and the IMDG Code; \hyperref[related-work]{Section 3} reviews related
work; \hyperref[data-description-and-methodology]{Section 4} describes the data sources, benchmark design, and
evaluation methodology; \hyperref[results]{Section 5} presents results; \hyperref[discussion]{Section 6} discusses
findings, limitations, and practical implications; \hyperref[conclusion]{Section 7} concludes.

\hypertarget{background}{%
\section{2. Background}\label{background}}

\hypertarget{the-imdg-code}{%
\subsection{2.1 The IMDG Code}\label{the-imdg-code}}

The International Maritime Dangerous Goods Code is the primary
international regulatory framework governing the transport of dangerous
materials by sea. Published and maintained by the International Maritime Organization (IMO), the Code
implements relevant provisions of the SOLAS (Safety Of Life At Sea) and MARPOL (Marine Pollutant) conventions
~\citep{imo_imdg_2024, joseph2021solas}, establishing a uniform framework
for the classification, documentation, packaging, marking, labelling, placarding, consignment, segregation,
stowage and handling of dangerous goods and environmentally hazardous substances.
It carries the force of international law across all signatory states
and applies to every sea shipment containing dangerous goods, regardless
of vessel size or trade route.
This global applicability, in contrast to national regulations
such as the US Code of Federal Regulations (CFR) Title 49,
makes the IMDG Code the operative standard for the vast majority of
containerised DG trade and is the focus of this benchmark.

The Code is updated on a two-year cycle with each amendment being valid
for three years \citep{HazcheckAmendment}. 
Each amendment is authorised for use in its first year, becomes mandatory
in its second and remains valid in its third, when the subsequent amendment may be applied voluntarily.
This gives a year-long overlap period where two versions of the Code are valid.
The current operative version, A42-24, was published in 2024 and took effect on 1
January 2025, with mandatory compliance from 1 January 2026. Substantive
changes introduced in A42-24 relative to A41-22 include an updated
definition for recycled plastics material, revised exemptions for carbon
and activated carbon, and extension of lithium battery marking
requirements to cover sodium-ion battery entries \citep{Hazcheck_A42}. The
next amendment, A43-26, will take effect from 1 January 2027,
at which point A42-24 becomes optional.
All questions in DGEval are grounded in A42-24.

Two areas of the Code are particularly relevant to this benchmark's results: stowage and segregation.
Stowage governs where on a vessel dangerous goods may be placed.
Each substance is assigned a stowage category (Categories A-E or 01-05 depending on the hazard class) specifying whether it may be carried on deck, under deck, or either, and may carry additional stowage codes imposing specific conditions such as keeping the goods away from living quarters, sources of heat or ignition, or other specified locations.
Segregation governs which dangerous goods must be kept physically apart from one another.
Incompatible substances, those that could interact dangerously in the event of leakage or fire, must be separated by minimum distances or intervening cargo and/or bulkheads, with requirements determined by the hazard classes involved and any segregation groups or codes assigned to the substances.

\hypertarget{the-dangerous-goods-list}{%
\subsection{2.2 The Dangerous Goods List}\label{the-dangerous-goods-list}}

The central reference table for DG practitioners is the Dangerous Goods
List\citep{imo_imdg_2024}. The DGL maps each
dangerous substance and article to its regulatory properties: a row is typically
identified by a United Nations number (UN No. a four-digit identifier assigned to a
specific dangerous good or class of goods) and a Proper Shipping Name
(PSN, the official transport name). The remaining 16 columns
include hazard class, packing group, segregation codes,
stowage requirements, emergency schedules, and more.
Two columns are duplicates, one is reserved, and several more contain multiple pieces of information.
For example the ``Marine Pollutant'' designation (P) is included in the ``Subsidiary Hazards'' column (column 4).
For the purposes of this paper, duplicate and reserved columns have been ignored and
columns containing multiple pieces of information have been separated out.
See \hyperlink{tab:a2}{Table A2} for the source DGL columns for each of the properties used.

These properties are often interdependent and correct interpretation of the DGL
frequently requires multi-step reasoning. For example, segregation requirements depend on hazard class,
subsidiary hazards and any additional requirements in column 16b; marking and placarding
requirements combine classification, packaging, and regulatory
conditions; all packaging, segregation and stowage decisions may be affected by exceptions
and additional requirements in columns 6 and 7, and by various provisions throughout the Code.
This interdependence introduces significant complexity: correct DGL
interpretation requires more than a simple table lookup, instead demanding
a thorough review of all applicable requirements, exceptions, and additional
provisions to ensure proper compliance.
See \hyperref[appendix-c-dgl-column-descriptions-and-weights]{Appendix C} for descriptions and expert-assigned weights for
all DGL fields evaluated in this benchmark.

\hypertarget{practitioner-roles}{%
\subsection{2.3 Practitioner Roles}\label{practitioner-roles}}

DG compliance spans several distinct practitioner roles, each with
different informational needs and different tolerances for error.
As well as Standard and Advanced courses, NCB Hazcheck's 
training taxonomy identifies four primary roles for testing:

\begin{itemize}
\tightlist
\item
  \textbf{Consignor/Freight Forwarder}: Personnel with management,
  supervisory, and safety compliance responsibilities for companies
  involved in DG.
\item
  \textbf{Packer/Cargo Handler}: Personnel who mark, label and placard dangerous goods packages and/or container transport units, including those who pack and unpack cargo transport units.
\item
  \textbf{Ship Loader}: Personnel responsible for or directly involved
  in ship loading and unloading operations.
\item
  \textbf{Ship Operator}: Personnel responsible for managing the
  operation of a vessel carrying dangerous goods.
\end{itemize}

These roles differ in which sections of the IMDG Code are most relevant
to their day-to-day work (see \hyperref[appendix-b-e-learning-subsection-details]{Appendix B}).

These roles are not independent. They are sequential steps in a single
transport chain, and a decision made at an early step constrains those that
follow. Segregation is the clearest example: the requirement to keep
incompatible goods apart is fixed when a container is packed, and it then
governs terminal handling, vessel loading, onboard stowage, and discharge. An
error made early, such as packing incompatible substances together, carries
through the rest of the chain and raises the risk faced by every later
participant. Compliance is therefore a property of the whole chain, not of any
single check, which is why model performance is worth examining role by role
rather than in aggregate (\hyperref[practitioner-analysis]{Section 5.3}). A
tool consulted at one step, human or automated, sees only that step, not the
decisions upstream or down.

\hypertarget{related-work}{%
\section{3. Related Work}\label{related-work}}

\subsection{3.1 Dangerous Goods Compliance and Regulatory Decision-Making}\label{related-work-dg-compliance}

Traditional dangerous goods compliance relies heavily on human expertise and manual consultation of the IMDG Code, often aided by digital tools. The Code mandates initial and recurrent training for personnel involved in the transport of dangerous goods~\citep{imo_imdg_2024}, reinforced by the parallel competence requirements of the STCW Convention~\citep{yi2025stcw}, and such expertise can be acquired through direct study of the Code and its supplements or dedicated e-learning courses, such as those offered by NCB Hazcheck. However, reliance on human judgement scales poorly and provides limited protection against error: a single overlooked provision can leave a shipment non-compliant, and incidents involving packaged dangerous goods account for an estimated 15\% of all container-ship casualty fatalities~\citep{ellis2011analysis}, with cargo fires a recurrent and difficult-to-contain consequence of such failures~\citep{kim2023roro}. Empirical studies have characterised port accidents involving chemical substances~\citep{lecue2019accidents}.

Automated systems have been developed to support and validate dangerous goods compliance. In commercial practice, the dominant approach encodes the Code's provisions as deterministic rule-based engines that validate classification, segregation, and documentation for a given consignment, and such tools are widely deployed across the container shipping industry~\citep{ncb_hazcheck}. Academic work has also focused on formalising the Code's knowledge to enable automated reasoning. One such approach represents substance properties as entities in an ontology and encodes the more complex stowage and segregation provisions as logical rules, so that segregation requirements between substances can be inferred automatically rather than consulted by hand~\citep{zhang2019construction}. The same segregation rules have also been incorporated into operations research models for container ship stowage planning, where they are used as constraints governing the feasible assignment of dangerous containers to stowage locations~\citep{pacino2011fast, ambrosino2015using}. These approaches improve the consistency and traceability of compliance checking but rely on manually engineered rules and ontologies that need to be updated with each amendment to the Code, and they remain limited to structured predefined queries rather than the open-ended natural language queries used in practice. Beyond consignment-level validation, regulatory compliance is also monitored at the enforcement level through port state control, where recent analyses increasingly draw on statistical and machine-learning methods to target inspection effort~\citep{sheriff2025psc}.

\subsection{3.2 Capabilities and Limitations of Large Language Models}

Large language models are AI systems trained on very large collections of text, spanning books, web pages, scientific literature, and technical documents, and optimised to predict the most plausible continuation of any piece of text. This training produces systems capable of answering open-ended questions across a wide range of subjects without being programmed with explicit rules for each domain~\citep{ouyang2022training}. Unlike the deterministic rule-based tools described in \hyperref[related-work-dg-compliance]{Section 3.1}, a single general-purpose LLM can respond to queries phrased in the way a practitioner would naturally ask them, without requiring a predefined query structure.

After initial training, models are refined to follow instructions and to align their outputs with human preferences~\citep{ouyang2022training}. Two additional techniques are directly relevant to this paper's evaluation. First, web search integration: when a model has web search enabled, it retrieves relevant pages from the internet at query time and incorporates them into its response, rather than relying solely on its training data~\citep{lewis2020retrieval}; this is tested in \hyperref[web-search]{Section 5.5}. Second, chain-of-thought reasoning: some configurations allocate more computation per query, allowing the model to work through a problem step by step before producing a final answer, which improves performance on complex, multi-step problems~\citep{wei2022chain}; this corresponds to the thinking levels evaluated throughout \hyperref[results]{Section 5}. These capabilities have made LLMs increasingly viable for knowledge-intensive tasks requiring factual accuracy and domain-specific reasoning.

LLMs nevertheless exhibit limitations that are consequential in safety-critical applications. In particular, they are prone to hallucination: generating fluent and plausible text that is factually incorrect or unsupported~\citep{huang2025survey}. Unlike a database lookup or a rule-based compliance engine, an LLM has no reliable internal signal for when it is uncertain; it produces a confident-sounding answer regardless of whether the information is present in its training data or not. Their behaviour is also sensitive to how a query is phrased, so that semantically equivalent questions can yield materially different answers~\citep{sclar2024quantifying}. Furthermore, performance on tasks requiring precise, multi-step reasoning over several interacting constraints remains unreliable, with accuracy degrading as problems grow in length or complexity~\citep{dziri2023faith}. Even retrieving external sources at query time, which mitigates some of these issues by grounding responses in authoritative text, does not eliminate them, as the model may still misread, ignore, or contradict the retrieved evidence~\citep{niu2024ragtruth}. These weaknesses are especially problematic in regulated domains such as dangerous goods compliance, where an answer must be accurate and traceable to an authoritative source, not merely fluent. Rigorous evaluation of LLMs in regulated, safety-critical domains is therefore essential.

\subsection{3.3 Safety-Critical and Domain-Specific Benchmarking of LLMs}

As large language models have grown more capable, standardised benchmarks have been proposed to quantify LLM performance and track progress. Early work focused on broad measures of knowledge and reasoning, such as MMLU~\citep{hendrycks2020measuring}, aggregating large numbers of tasks to assess general competence, alongside capability-specific benchmarks such as GPQA for graduate-level scientific reasoning~\citep{rein2023gpqagraduatelevelgoogleproofqa}. As leading models increasingly saturate many of these benchmarks, focus has shifted to harder and more discriminating evaluations, including MMLU-Pro~\citep{wang2024mmluprorobustchallengingmultitask} and Humanity's Last Exam~\citep{phan2025humanity}. Strong performance on these general-purpose benchmarks, however, does not reliably transfer to specialised domains, motivating domain-specific benchmark development.

Domain-specific benchmarks have been developed to evaluate LLM performance on tasks requiring specialised knowledge and reasoning over authoritative materials. In the legal domain, benchmarks such as LegalBench~\citep{guha2023legalbenchcollaborativelybuiltbenchmark} assess statutory interpretation and legal reasoning, while FinanceBench~\citep{islam2023financebenchnewbenchmarkfinancial} evaluates financial reasoning grounded in reports and regulatory disclosures. More recent benchmarks have moved beyond answer correctness alone to assess whether responses are grounded in supporting evidence, either through explicit evaluation of retrieval-augmented systems, as in LegalBench-RAG~\citep{pipitone2024legalbench}, or through expert-authored rubrics for grading open-ended responses, as in HealthBench~\citep{arora2025healthbench}. General-purpose evaluations do not reliably predict domain-specific performance; these benchmarks confirm the gap.

The case for specialised evaluation is strongest in safety-critical domains, where errors have direct operational consequences. LabSafety Bench evaluates hazard identification, risk assessment and consequence prediction in scientific laboratories and reports substantial performance limitations among existing models, with no evaluated system exceeding 70\% accuracy on hazard identification~\citep{Zhou2026}. A recent review of LLM benchmarking for safety-critical hazard analysis identifies persistent gaps in causal reasoning evaluation and the absence of any assessment of regulatory compliance, and it calls for specialised benchmarks to systematically evaluate LLM capabilities before their deployment in safety-critical applications~\citep{DOKAS2026107056}.
Dangerous goods compliance is precisely this kind of domain: practitioners must identify hazards, interpret regulatory requirements, and reason across interacting constraints governing packaging, segregation, and stowage. Errors in these decisions have been linked to faults during cargo preparation, packing and loading, which remain among the leading causes of dangerous goods incidents at sea~\citep{ellis2011analysis}.

Despite the growing number of domain-specific and safety-critical benchmarks, none evaluates dangerous goods compliance. While LLaMarine~\citep{nguyen2025llamarineopensourcemaritimeindustryspecific} adapts language models to the maritime domain, it is not an evaluation benchmark and does not address dangerous goods compliance. To address this gap, we introduce DGEval, a specialised benchmark for evaluating large language models on dangerous goods compliance tasks grounded in the IMDG Code. It is precisely the kind of compliance-focused, safety-critical benchmark that this literature calls for, and it fills both the specific absence of a dangerous goods benchmark and the wider gap in regulatory-compliance evaluation.

\hypertarget{data-description-and-methodology}{%
\section{4. Data Description and
Methodology}\label{data-description-and-methodology}}

\hypertarget{data-sources}{%
\subsection{4.1 Data Sources}\label{data-sources}}

DGEval draws on two primary proprietary data sources.

\textbf{NCB Hazcheck e-learning platform.} The NCB Hazcheck platform contains
many questions developed by DG subject-matter experts for practitioner
training and certification across the four practitioner roles described
in \hyperref[practitioner-roles]{Section 2.3}. Of these, 609 questions are aligned with IMDG A42-24
and span 20 subsections covering topics from general provisions to
classification, packing, consignment, stowage, segregation, and
emergency response. Questions containing image references (e.g., ``which
of these images is a marine pollutant placard?'') were excluded, leaving
520 usable questions.
The great majority (\textgreater90\%) offer four
answer options with one correct answer; a small proportion offer up to
eight options with up to four correct answers. Detailed subsection
descriptions and expert-assigned importance weights are provided in
\hyperref[appendix-b-e-learning-subsection-details]{Appendix B} and \hyperlink{tab:a1}{Table A1}. 

\textbf{Dangerous Goods List.} 

NCB Hazcheck’s own curated DGL, aligned with IMDG Amendment 42-24~\citep{imo_imdg_code_2024}, was used to construct structured lookup questions.\footnote{No extracts from the published IMDG Code were provided to the evaluated models as source material.}
In consultation with in-house DG experts, we assigned importance weights to
16 DGL columns (excluding UN number and PSN, which were weighted
separately at 25\% each). UN/PSN combinations with variant entries were
excluded to avoid ambiguity. This process yielded 9,443 eligible questions across three question types:

\begin{itemize}
\tightlist
\item
  Given a UN number, what is the PSN? (2,347 questions)
\item
  Given a PSN, what are the possible UN numbers? (2,402 questions)
\item
  Given a UN number and PSN, what is the value of a specific DGL
  property? (4,694 questions)
\end{itemize}

A random subsample of 22 questions was reviewed by in-house DG experts
to confirm correctness.

The benchmark questions and answer key are not released publicly. This is a deliberate design choice, not a limitation. If both were public, future models could simply be trained on the test set, and a high score would show memorisation of Q\&A pairs rather than understanding of the IMDG Code. DGEval instead follows the controlled-release model used by established benchmarks such as nuScenes~\citep{caesar2020nuscenes}, where participants submit predictions to an evaluation server and receive official scores without direct access to the hidden test labels. A submission-based evaluation server following this model is planned; see Appendix E.

\hypertarget{benchmark-sections}{%
\subsection{4.2 Benchmark Sections}\label{benchmark-sections}}

DGEval comprises four sections designed to probe distinct types of DG
practitioner knowledge. \hyperlink{tab:1}{Table 1} summarises the composition of each
section.

\hypertarget{tab:1}{}%
\textbf{Table 1: Benchmark composition.}

\begin{longtable}[]{@{}lllll@{}}
\toprule\noalign{}
Section & Name & Source & Question Type & Questions \\
\midrule\noalign{}
\endhead
\bottomrule\noalign{}
\endlastfoot
1 & MCQ Q\&A & E-learning & Multiple choice (closed book) & 520 \\
2 & Open-Ended Q\&A & E-learning (adapted) & Free text (LLM-as-judge) & 360 \\
3 & DGL Lookup & DGL & Structured lookup & 485 \\
4 & Regulatory Recall & E-learning (derived) & IMDG section identification & 313 \\
\textbf{Total} & & & & \textbf{1,678} \\
\end{longtable}

\textbf{Section 1 - Multiple-Choice Q\&A.} All 520 non-image e-learning questions
are presented to models exactly as they appear on the NCB Hazcheck platform:
multiple-choice, closed-book, with the same answer options provided to
human candidates.
By posing the exact same questions as faced by human candidates, we can directly compare model performance with a human baseline (see \hyperref[human-baseline]{4.4}).

\smallskip
\noindent\textit{Example, Question E53/4:} ``What is the objective of
the UN packaging specification scheme?''
\begin{itemize}
\tightlist
\item To prove the competence of the packaging design and materials of construction for carrying dangerous goods \textbf{[Correct]}
\item To prove which dangerous goods are compatible with the different packaging types
\item To prove that dangerous goods in ``limited quantities'' do not need to be carried in approved packagings
\item To confirm the type of packagings permitted for a particular substance
\end{itemize}

\textbf{Section 2 - Open-Ended Q\&A.} The same e-learning questions are
reformulated without the multiple-choice answer options, requiring
models to produce a free-text answer.
This tests a qualitatively different capability from Section 1: providing answer options gives substantial scaffolding, allowing a model to succeed by recognising a correct answer rather than generating one.
Practitioners consulting an LLM in practice will typically pose open questions rather than multiple-choice queries, so the open-ended format is more representative of real deployment conditions and provides a stronger test of whether a model holds the underlying knowledge.

Questions whose phrasing only makes sense in the presence of answer options (e.g. ``Which of these\ldots'' or ``Which of the following\ldots'') were removed, reducing the 520 e-learning questions to 360 usable items.
For questions with multiple correct answers, models were expected to enumerate all correct answers, with partial credit applied where only a subset was identified.

Automated evaluation is performed using an LLM as a judge. The LLM-as-judge
paradigm has been widely adopted for evaluating free-text outputs at
scale \citep{electronics15030659}.
We follow this methodology with expert validation, as detailed in \hyperref[scoring]{Section 4.3} and \hyperref[appendix-d-llm-judge-validation]{Appendix D}.

\textbf{Section 3 - DGL Lookup.} Using the importance weights assigned
by DG experts, 485 questions were randomly sampled from the 9,443
eligible questions, with UN number and PSN queries each comprising
approximately 25\% of the sample. DGL questions appearing in the
e-learning dataset were excluded to prevent overlap with Sections 1 and
2.

A secondary motivation for Section 3 was the prevalence of DGL-related
questions within the e-learning dataset itself. Of the 520 questions in
Section 1, 82 (15.8\%) involve querying the DGL, spread across multiple
subsections, with the largest concentrations in E52 (Classification) and
E61 (Additional Classification Considerations). This proportion is
substantial enough to warrant dedicated analysis, but within the mixed
Section 1 results the DGL signal is confounded by other question types.
Section 3 was therefore designed to isolate DGL retrieval as a distinct
competency, using a larger and more systematically sampled set of 485
questions that give finer-grained visibility into which specific DGL
properties models can reliably retrieve.

\smallskip
\noindent\textit{Example, Question 1:} ``What is the Proper Shipping Name of UN0004?''

\noindent\textit{Answer:} AMMONIUM PICRATE

\textbf{Section 4 - Regulatory Recall.} E-learning multiple-choice
questions were filtered to retain only those for which a specific IMDG
Code section had been identified by experts. Additionally, DGL-related
questions (all drawn from Section 3.2.1 of the Code) and those
containing an IMDG Code section were excluded. Multiple-choice answers
were provided, unless any one of them contained IMDG section codes; then
no MCQ answers were given. For each remaining question, 313 in total,
models were prompted to identify the specific chapter(s) and section(s)
of the IMDG Code that contain the answer, rather than to answer the
question itself.

\begin{figure}[htbp]
  \centering
  \includegraphics[width=\textwidth]{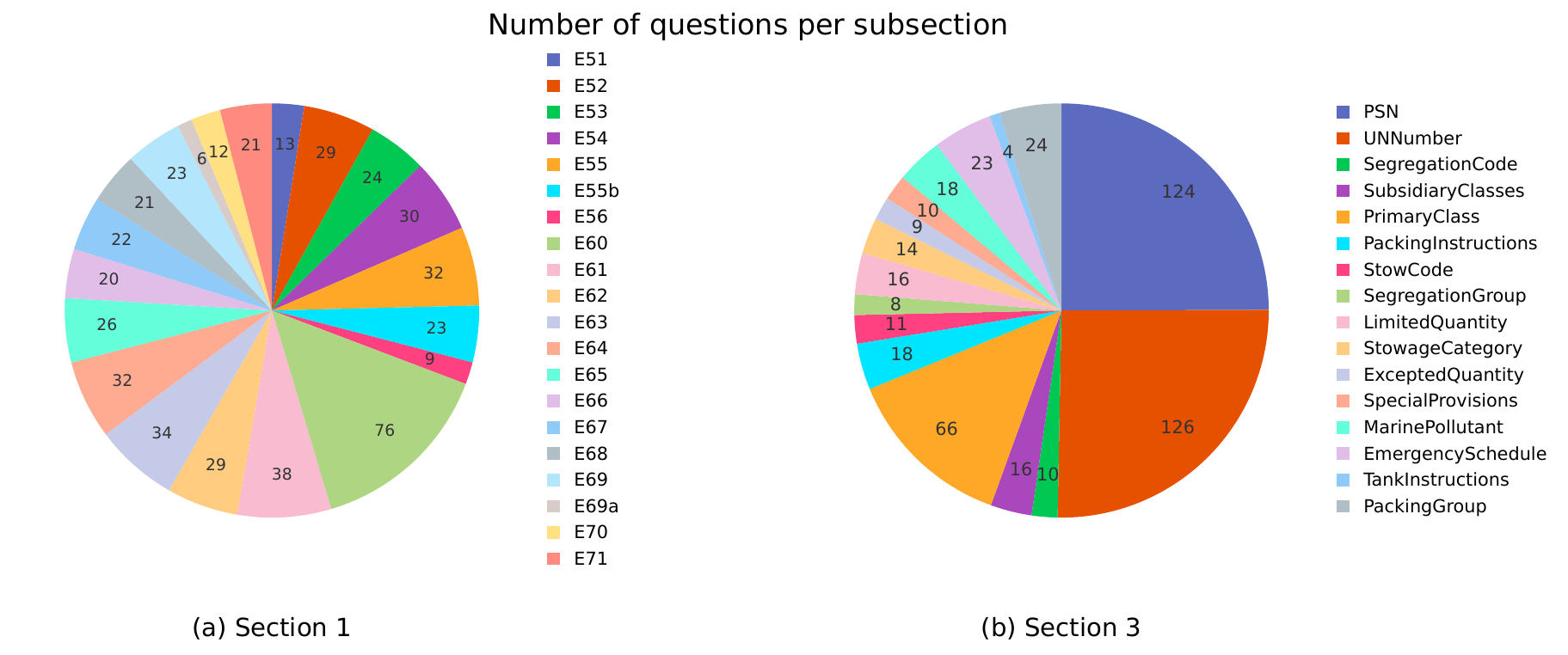}
  \caption{Question counts per e-learning subsection across benchmark sections.}
  \label{fig:subsection-counts}
\end{figure}

Sections 1, 2, and 3 measure the ability to recall DG knowledge, though that knowledge need not have come from the IMDG Code directly.
Section 4 tests direct understanding of the Code's structure: which chapter or section contains the answer, giving practitioners a route to the authoritative source.
Figure~\ref{fig:subsection-counts} shows how the e-learning questions are distributed across subsections and across the benchmark sections drawn from them.

\hypertarget{scoring}{%
\subsection{4.3 Scoring}\label{scoring}}

\textbf{Sections 1, 3, and 4 (MCQ, Open-ended Q\&A and lookup).} For questions with a
single correct answer, a binary correct/incorrect score is assigned. For
questions with multiple correct answers (e.g., ``What are the subsidiary classes for UN1295, "TRICHLOROSILANE"?''), a partial credit scheme is applied: if
there are \(n\) correct answers and the model identifies \(k\) of them,
the score is \(k/n\). If the model provides more answers than exist, a
symmetric penalty is applied (See \hyperref[appendix-a-evaluation-metrics-detail]{Appendix A}). In practice, extracting model answers
required section-specific regular expression parsing.

\textbf{Section 2 (open-ended).} Manual evaluation at the scale of
Section 2 is impractical, so we employ an LLM judge. We used
gemini-2.5-flash (non-thinking) as the judge model. The judge
was prompted to classify each model response as correct,
partial, or incorrect, scored as 1.0, 0.5, and 0.0
respectively. A small proportion (0.08\%) of responses received an
unknown judgment and were discarded. Judge validity was
assessed by human expert review of a sample of responses; details are
provided in \hyperref[appendix-d-llm-judge-validation]{Appendix D}. Because the judge shares a provider with the
top-scoring model (Gemini 3.1 Pro), self-preference bias is possible, a
recognised effect whereby a model judge may favour outputs from its own
family \citep{panickssery2024llmevaluators}. This expert validation indicates that the judge tracks human
grading closely, which limits this concern, although we did not test for
family-specific favouritism directly; a cross-family or ensemble judge
would remove it and is left to future work.

\textbf{Subsection weighting.} Within Sections 1, 2, and 4, questions
are grouped by e-learning subsection (E51-E71). Subsection scores are calculated as an equal-weighted average of the questions within that subsection. Two in-house NCB Hazcheck
DG experts independently rated each subsection on a 1-10 scale of
danger-criticality; inter-rater disagreements were discussed until
consensus was reached. Section scores are then calculated as a weighted average of the subsection scores using normalised weights. See \hyperlink{tab:a1}{Table A1} in \hyperref[appendix-b-e-learning-subsection-details]{Appendix B} for subsection weights.

Within Section 3, the same process was
undertaken, with the subsections being given by the DGL column. Scores
are similarly aggregated using the column importance weights (\hyperlink{tab:a2}{Table A2}, \hyperref[appendix-c-dgl-column-descriptions-and-weights]{Appendix C}).

\textbf{Section-level scores and composite.} The composite-level scores were the average of the four section scores with equal weighting.

\hypertarget{human-baseline}{%
\subsection{4.4 Human Baseline}\label{human-baseline}}

The NCB Hazcheck e-learning platform records candidate performance on the
e-learning assessments. The platform pass mark is 75\%, and, to date, the raw
average across all candidate attempts is 88.4\%. However, this figure
overstates single-attempt performance because candidates take assessments immediately after viewing the course material, the platform allows unlimited attempts and, most importantly, scores below 75\% are not recorded.

To correct for the unlimited attempts, we retained only the
first attempt per candidate per assessment element, averaged per-element
scores within each candidate, and then averaged across candidates. After
this adjustment, the singly-corrected human baseline on Section 1 questions was
87.7\%.

To estimate the true population mean of the test scores, the data was then adjusted for two distinct distributional anomalies: a strict lower recording threshold at $75\%$ (left-truncation) and a pronounced ceiling effect, where a substantial proportion of passing students ($\approx 11.1\%$) scored exactly 100\% \citep{OSPINA20121609}. Simple sample mean calculations of the observed data thus yield a severely upward-biased estimate. To resolve this, we employed a two-part mixture model approach. The discrete probability mass at the 100\% boundary was isolated and treated as a distinct sub-population. The remaining continuous data spanning the $[0.75, 1.0)$ interval was modelled using a left-truncated Beta distribution, which natively accounts for both the lower truncation threshold and the bounded $[0, 1]$ nature of percentage-based scores. The shape parameters ($\alpha = 8.91$, $\beta = 1.93$) of the underlying distribution were estimated via maximum likelihood estimation (MLE). By mathematically reconstructing the latent probability density function below the 75\% threshold (estimated at 21.37\% of the total population), we then derived a corrected, weighted population mean of 83.8\%. This figure depends on the assumed Beta form and the fitted parameters used to reconstruct the scores below the 75\% threshold, and should therefore be read as an approximate, model-dependent estimate.

We remain unable to adjust for the fact that the human candidates take the e-learning tests immediately after observing the course material. However, this recency effect is a systemic feature of the dataset, and we are mindful that we must treat this inferred population mean as an upper bound.

Two further caveats apply to the interpretation of this baseline. First, we do not know the level of prior DG expertise held by the e-learning candidates, who range from newcomers seeking initial certification to experienced practitioners undertaking recurrent training; the baseline therefore reflects a mixed population rather than a defined level of competence. Second, the comparison measures recall alone. In practice, a practitioner establishing compliance would consult the IMDG Code itself or a digital compliance tool rather than work from memory, so real-world accuracy should be considerably higher than this closed-book figure; indeed, in an operational context anything short of 100\% represents a regulatory failure. This baseline should not, therefore, be read as an estimate of achievable real-world compliance accuracy. Because both the human candidates and the LLMs are assessed here under the same closed-book, recall-based conditions, however, the comparison between them remains like-for-like.

\hypertarget{model-selection}{%
\subsection{4.5 Model Selection}\label{model-selection}}

We evaluated 13 models spanning six providers and families:

\textbf{Commercial models:}
\begin{itemize}
\tightlist
\item \emph{Anthropic}: Claude Sonnet 4.6 (small) and Claude Opus 4.7 (large)
\item \emph{Google}: Gemini 3.1 Flash-Lite (small) and Gemini 3.1 Pro (large)
\item \emph{OpenAI}: GPT-5.4 Mini (small) and GPT-5.5 (large)
\end{itemize}

\textbf{Open-source models} (hosted locally on dual 96 GB GPU machine):
\begin{itemize}
\tightlist
\item \emph{Google}: Gemma 4 (4B, small) and Gemma 4 (31B, large)
\item \emph{Meta}: Llama 3.2 (3B, small) and Llama 3.3 (70B, large)
\item \emph{Qwen}: Qwen3.5 (9B, small) and Qwen3.6 (35B, large)
\end{itemize}

\textbf{Domain-specific:}
\begin{itemize}
\tightlist
\item \emph{LLaMarine}: Llama 3.1 (70B) fine-tuned on maritime regulatory text \citep{nguyen2025llamarineopensourcemaritimeindustryspecific}
\end{itemize}

Within each commercial model, we tested multiple thinking levels; where additional compute resources are used so that the model can generate a chain-of-thought (CoT), in which a complex problem is decomposed into a series of intermediate reasoning steps before a final answer is generated.
While the specific implementations of this feature differ greatly by provider, to the typical DG practitioner they are functionally the same: higher thinking levels permit the model to use more tokens before reaching an answer.
Commercial models were tested at up to five thinking levels (None, Low, Medium, High, Max for Claude; Minimal, Low, Medium, High for Gemini; None, Low, Medium, High for GPT).
Open-source models were tested at None and Thinking (where supported).

All models were accessed programmatically via API between April and May 2026. Commercial
models were accessed via vendor APIs; open-source models were hosted
locally on a dedicated GPU server and accessed using the OpenAI-compatible API.
All models had web search
disabled by default (see \hyperref[web-search]{Section 5.5} for a targeted web-search
sub-analysis). Cost data were only collected for commercial models.
Time-taken data was collected for all models.
Cost and latency are reported because DG compliance is high-volume work in
which deployment is constrained by budget and turnaround, not accuracy alone.
Each question was evaluated once per model configuration; because several
models are non-deterministic, the reported scores are single-run point
estimates and do not capture run-to-run variance.

\hypertarget{results}{%
\section{5. Results}\label{results}}

\hypertarget{overall-performance}{%
\subsection{5.1 Overall Performance}\label{overall-performance}}

\begin{figure}[htbp]
  \centering
  \includegraphics[width=\textwidth]{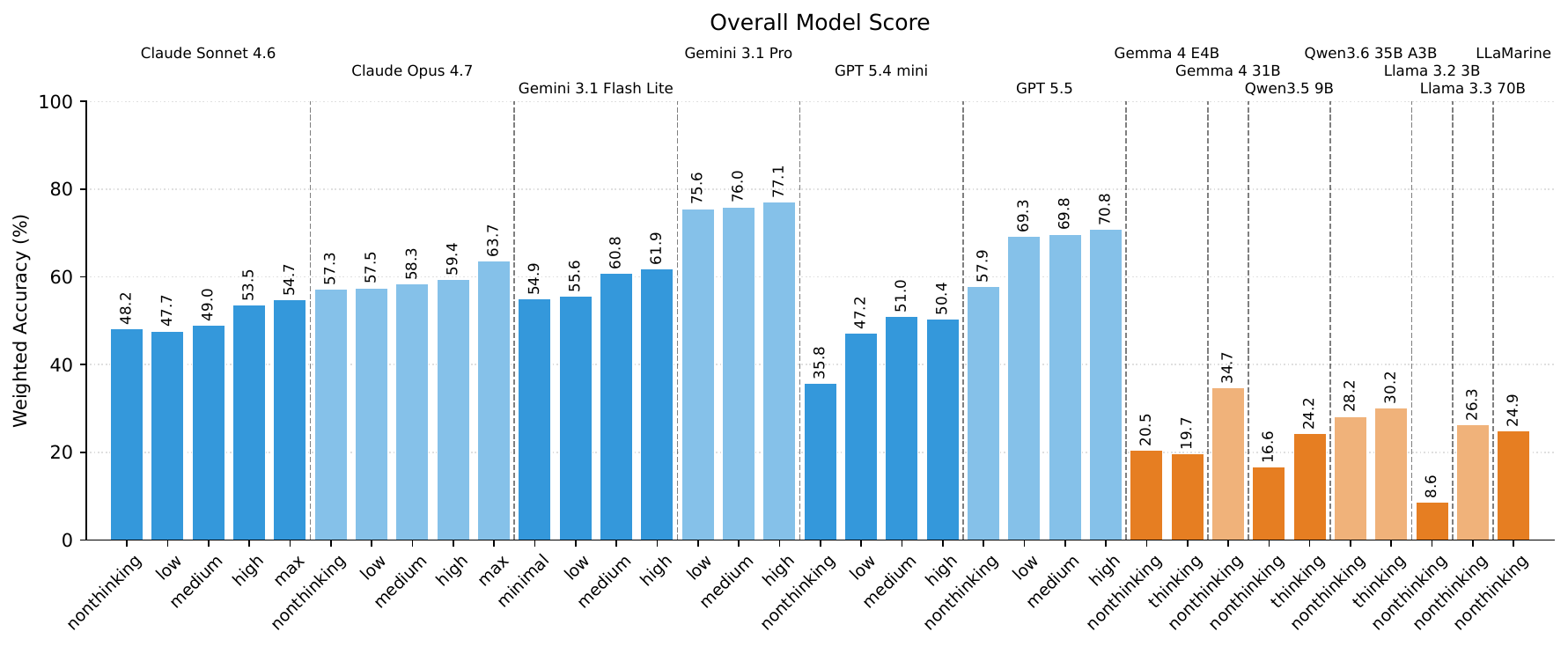}
  \caption{Total weighted benchmark scores for all model configurations.}
  \label{fig:total-weighted-score}
\end{figure}

Figure~\ref{fig:total-weighted-score} plots the total weighted benchmark score for every model configuration, and \hyperlink{tab:2}{Table 2} reports the full per-section breakdown.
Gemini 3.1 Pro leads all other tested models across all four
sections. At its highest thinking level, it achieves 95.8\%, 86.0\%,
89.6\%, and 36.8\% on Sections 1-4 respectively, with no other model
coming within 5 percentage points on Section 1. Sections 1, 2, and 3
show broadly comparable section means (65.7\%, 59.1\%, and 54.9\%
respectively across all configurations), while Section 4 is dramatically
harder, with a mean of only 14.2\%, indicating that identifying the
precise IMDG regulatory section containing an answer is a qualitatively
different and more demanding task than answering questions about it.

On Section 1, ten model configurations exceed the 75\% practitioner pass
mark: all three Gemini 3.1 Pro configurations, GPT-5.5 at
Low/Medium/High, GPT-5.4 Mini at High, Gemini Flash-Lite at Medium and
High, and Claude Sonnet 4.6 at Max. Only Gemini 3.1 Pro and GPT-5.5 (at
Low and above) exceed the corrected human baseline of 83.8\%.

The commercial/open-source gap is substantial. The best open-source
result on Section 1 is Gemma 4 (31B) at 66.2\%, compared to GPT-5.4 Mini
(None) at 53.6\% as the weakest commercial model. This gap widens
sharply on Section 3: Gemma 4 (31B) scores only 23.3\% on DGL lookup,
while GPT-5.4 Mini (None) scores 32.2\%, and the best commercial model
(Gemini 3.1 Pro High) scores 89.6\%. On Section 4, the gap is even
starker: the best open-source score is Llama 3.3 (70B) at 6.6\%, while
most commercial models at low thinking exceed 20\%.

\hypertarget{tab:2}{}%
\textbf{Table 2: DGEval results - all model configurations.} Human
practitioner baseline on Section 1 is 83.8\%. S1: MCQ; S2: Open-ended;
S3: DGL Lookup; S4: Regulatory Recall.
Bold indicates section-level maximum.

\begingroup\footnotesize
\begin{longtable}[]{@{}llllrrrr@{}}
	\toprule\noalign{}
	Model & Provider & Size & Thinking & S1 (\%) & S2 (\%) & S3 (\%) & S4 (\%) \\
	\midrule\noalign{}
	\endhead
	\bottomrule\noalign{}
	\endlastfoot
	Claude Sonnet 4.6 & Anthropic & Small & None & 57.9 & 60.4 & 63.7 & 10.6 \\
	Claude Sonnet 4.6 & Anthropic & Small & Low & 59.6 & 57.8 & 64.3 & 9.0 \\
	Claude Sonnet 4.6 & Anthropic & Small & Medium & 59.9 & 60.8 & 64.0 & 11.4 \\
	Claude Sonnet 4.6 & Anthropic & Small & High & 67.6 & 65.3 & 73.6 & 7.7 \\
	Claude Sonnet 4.6 & Anthropic & Small & Max & 75.5 & 62.1 & 74.2 & 7.5 \\
	Claude Opus 4.7 & Anthropic & Large & None & 65.8 & 68.2 & 77.5 & 17.4 \\
	Claude Opus 4.7 & Anthropic & Large & Low & 68.3 & 66.9 & 78.6 & 16.2 \\
	Claude Opus 4.7 & Anthropic & Large & Medium & 67.6 & 70.1 & 78.0 & 17.6 \\
	Claude Opus 4.7 & Anthropic & Large & High & 67.8 & 74.7 & 77.9 & 17.2 \\
	Claude Opus 4.7 & Anthropic & Large & Max & 72.1 & 78.8 & 83.7 & 20.2 \\
	Gemini 3.1 Flash-Lite & Google & Small & Minimal & 65.9 & 67.5 & 67.9 & 17.8 \\
	Gemini 3.1 Flash-Lite & Google & Small & Low & 66.3 & 69.6 & 68.7 & 17.9 \\
	Gemini 3.1 Flash-Lite & Google & Small & Medium & 79.2 & 71.0 & 73.8 & 19.3 \\
	Gemini 3.1 Flash-Lite & Google & Small & High & 79.0 & 72.8 & 73.7 & 22.0 \\
	Gemini 3.1 Pro & Google & Large & Low & 94.4 & 84.3 & 89.4 & 34.1 \\
	Gemini 3.1 Pro & Google & Large & Medium & 94.0 & \textbf{86.5} & 88.4 & 35.0 \\
	Gemini 3.1 Pro & Google & Large & High & \textbf{95.8} & 86.0 & \textbf{89.6} & \textbf{36.8} \\
	GPT-5.4 Mini & OpenAI & Small & None & 53.6 & 48.4 & 32.2 & 8.8 \\
	GPT-5.4 Mini & OpenAI & Small & Low & 71.0 & 60.4 & 43.7 & 13.7 \\
	GPT-5.4 Mini & OpenAI & Small & Medium & 74.0 & 67.0 & 47.5 & 15.6 \\
	GPT-5.4 Mini & OpenAI & Small & High & 76.1 & 66.5 & 46.4 & 12.6 \\
	GPT-5.5 & OpenAI & Large & None & 71.9 & 68.4 & 69.2 & 22.0 \\
	GPT-5.5 & OpenAI & Large & Low & 90.1 & 80.0 & 81.2 & 25.8 \\
	GPT-5.5 & OpenAI & Large & Medium & 89.6 & 79.6 & 82.9 & 27.0 \\
	GPT-5.5 & OpenAI & Large & High & 90.7 & 79.2 & 82.7 & 30.5 \\
	Gemma 4 (4B) & Google & Small & None & 47.3 & 22.8 & 8.8 & 3.1 \\
	Gemma 4 (4B) & Google & Small & Thinking & 37.8 & 22.1 & 16.0 & 2.9 \\
	Gemma 4 (31B) & Google & Large & None & 66.2 & 44.1 & 23.3 & 5.2 \\
	Qwen3.5 (9B) & Qwen & Small & None & 32.2 & 25.8 & 8.0 & 0.5 \\
	Qwen3.5 (9B) & Qwen & Small & Thinking & 54.9 & 29.7 & 11.4 & 1.0 \\
	Qwen3.6 (35B) & Qwen & Large & None & 60.2 & 36.2 & 14.3 & 2.1 \\
	Qwen3.6 (35B) & Qwen & Large & Thinking & 59.5 & 41.9 & 17.7 & 1.6 \\
	Llama 3.2 (3B) & Meta & Small & None & 14.6 & 15.0 & 4.6 & 0.3 \\
	Llama 3.3 (70B) & Meta & Large & None & 36.3 & 39.2 & 23.0 & 6.6 \\
	LLaMarine & Fine-tuned & 70B & None & 35.7 & 40.7 & 20.0 & 3.4 \\
\end{longtable}
\endgroup

LLaMarine performs approximately at the level of base Llama 3.3 (70B)
across all sections despite being fine-tuned for maritime domain: 35.7\%
vs 36.3\% on Section 1, 40.7\% vs 39.2\% on Section 2, 20.0\% vs 23.0\%
on Section 3, and 3.4\% vs 6.6\% on Section 4. This is consistent with
LLaMarine's base model being older (Llama 3.1 (70B)), and trained on
broader maritime regulation rather than DG specifically.

\hypertarget{subsection-analysis}{%
\subsection{5.2 Subsection Analysis}\label{subsection-analysis}}

\hypertarget{section-1-mcq-by-subsection}{%
\subsubsection{Section 1 (MCQ) by
Subsection}\label{section-1-mcq-by-subsection}}

Section 1 shows a clear competency gradient across subsections.
Classification and procedural subsections are easiest: E60
(Classification - Procedures \& Code), E56 (Documentation), and E52
(Classification - Substances \& PSN) average 82-85\% across all model
configurations. Operational subsections are substantially harder: E65
(Stowage of Dangerous Goods) and E67 (Segregating Dangerous Goods on
Board Vessels) both average approximately 49-50\%, a gap of more than
35 percentage points. The spread across subsections is 35.8 percentage
points.

Even Gemini 3.1 Pro, the best-performing model overall, shows this
hierarchy: near-perfect performance on classification subsections, but
weaker performance on stowage and segregation, suggesting these are
genuine task-difficulty effects rather than model-specific limitations.

\hypertarget{section-2-open-ended}{%
	\subsubsection{Section 2 (Open-Ended)}\label{section-2-open-ended}}

Section 2 broadly mirrors the Section 1 subsection
hierarchy, with classification subsections easier and stowage/segregation
harder. The overall Section 2 mean across all configurations is 59.1\%,
about 6.5 percentage points below the Section 1 mean (65.7\%), consistent
with free-text generation being harder than selecting among provided options.
The gap between the two sections is more pronounced for open-source models:
Gemma 4 (31B) drops from 66.2\% on Section 1 to 44.1\% on Section 2 (a 22.1 pp gap),
and Qwen3.6 (35B) drops from 60.2\% to 36.2\% (24.0 pp).
Commercial models show smaller drops; Gemini 3.1 Pro Low falls only 10.1 pp (94.4\% to 84.3\%),
and several Claude configurations show gaps of 5 pp or less.
This suggests open-source models can identify correct answers when options are provided
but struggle to generate them without that scaffolding. Gemini 3.1 Pro remains the best
performer on Section 2, reaching 86.5\% at Medium thinking.

\hypertarget{section-3-dgl-lookup-by-field}{%
\subsubsection{Section 3 (DGL Lookup) by
Field}\label{section-3-dgl-lookup-by-field}}

Section 3 shows dramatic variation by field type, with a spread of 62.1
percentage points between the easiest and hardest fields (averaged
across all models). Three performance tiers are evident:

\textbf{Easy (mean \textgreater70\%):} Marine Pollutant (80.5\%),
Primary Hazard Class (74.3\%), Packing Group (71.0\%). These fields have clear,
unambiguous values for most substances and are frequently referenced in
general chemistry and safety literature.

\textbf{Moderate (mean 40-70\%):} Eight fields ranging from Emergency
Schedule (50.6\%) to PSN (44.9\%). These require more precise regulatory
knowledge.

\textbf{Hard (mean \textless40\%):} Five fields: Stowage Code (39.8\%),
Stowage Category (36.1\%), Segregation Code (25.5\%), Special Provisions
(21.3\%), and Segregation Group (18.4\%). These are almost entirely
operational and safety-critical fields that are unlikely to be
well-represented in general pre-training corpora. Gemini 3.1 Pro High
achieves 100\% on the easiest fields (Marine Pollutant, Primary Hazard Class,
UN Number) but only 51.5\% on Stowage Code and 79.2\% on Segregation Group,
which confirms these are task-difficulty effects rather than a
weakness specific to one model.

The weakest model, Llama 3.2 (3B), achieves 0\% on 12 of 16 fields, with
above-zero performance only on Marine Pollutant (61.1\%) and Primary Hazard 
Class (23.4\%). This indicates a qualitative failure on operational
field retrieval rather than simple performance degradation.

\hypertarget{section-4-reg-recall}{%
\subsubsection{Section 4 (Regulatory Recall)}\label{section-4-reg-recall}}

Section 4 is uniformly difficult across all subsections and models,
ranging from 4.8\% (E69, Transport Operations - Containers) to 24.2\% (E51, General Provisions)
when averaged across models. The relative ease of E51 may reflect that general provisions
are the most commonly cited section of the Code in derivative literature, giving models
incidental exposure to references to that section; operational subsections such as E66, E67,
and E69 cover vessel-specific procedures that are less likely to appear in pre-training sources.
Gemini 3.1 Pro reaches 36.8\% at High thinking, with GPT-5.5 at High (30.5\%) the only
other configuration above 30\%. Most commercial models cluster between 17\% and 27\%,
while open-source models are almost entirely below 7\%: Llama 3.3 (70B) at 6.6\% is the
best open-source result. Section 4 is also the section where Gemini 3.1 Pro's lead over
the rest of the commercial field is widest, with more than 6 percentage points separating
it from the next-best model, compared to narrower gaps on Sections 1, 2, and 3.

\hypertarget{practitioner-analysis}{%
\subsection{5.3 Practitioner Analysis}\label{practitioner-analysis}}

\hyperlink{tab:3}{Table 3} reports Section 1 scores broken down by the four practitioner roles.

\hypertarget{tab:3}{}%
\textbf{Table 3: Section 1 scores by practitioner role.} C/FF: Consignor/Freight Forwarder; P/CH: Packer/Cargo Handler; SL: Ship Loader; SO: Ship Operator. Bold indicates practitioner-level maximum.

\begingroup\footnotesize
\begin{longtable}[]{@{}llrrrr@{}}
	\toprule\noalign{}
	Model & Thinking & C/FF (\%) & P/CH (\%) & SL (\%) & SO (\%) \\
	\midrule\noalign{}
	\endhead
	\bottomrule\noalign{}
	\endlastfoot
	Claude Sonnet 4.6 & None & 67.8 & 69.7 & 52.2 & 54.8 \\
	Claude Sonnet 4.6 & Low & 71.9 & 72.0 & 52.7 & 55.5 \\
	Claude Sonnet 4.6 & Medium & 72.8 & 71.0 & 53.5 & 58.3 \\
	Claude Sonnet 4.6 & High & 81.1 & 82.2 & 64.3 & 65.8 \\
	Claude Sonnet 4.6 & Max & 84.4 & 82.3 & 70.8 & 73.5 \\
	Claude Opus 4.7 & None & 77.4 & 80.3 & 61.2 & 62.5 \\
	Claude Opus 4.7 & Low & 77.8 & 81.3 & 65.7 & 65.9 \\
	Claude Opus 4.7 & Medium & 76.3 & 78.6 & 63.6 & 64.2 \\
	Claude Opus 4.7 & High & 77.6 & 83.8 & 69.1 & 64.7 \\
	Claude Opus 4.7 & Max & 81.7 & 85.4 & 71.7 & 71.0 \\
	Gemini 3.1 Flash-Lite & Minimal & 77.7 & 82.1 & 60.7 & 61.7 \\
	Gemini 3.1 Flash-Lite & Low & 75.9 & 77.7 & 65.3 & 64.6 \\
	Gemini 3.1 Flash-Lite & Medium & 84.0 & 89.1 & 77.8 & 78.2 \\
	Gemini 3.1 Flash-Lite & High & 85.5 & 90.3 & 78.7 & 75.2 \\
	Gemini 3.1 Pro & Low & 96.1 & \textbf{100.0} & 92.6 & 93.6 \\
	Gemini 3.1 Pro & Medium & 94.6 & 98.8 & 92.1 & 93.5 \\
	Gemini 3.1 Pro & High & \textbf{96.3} & 98.8 & \textbf{94.8} & \textbf{95.3} \\
	GPT-5.4 Mini & None & 54.8 & 56.7 & 48.4 & 52.9 \\
	GPT-5.4 Mini & Low & 72.0 & 76.2 & 67.7 & 66.2 \\
	GPT-5.4 Mini & Medium & 79.1 & 84.5 & 72.0 & 67.7 \\
	GPT-5.4 Mini & High & 80.7 & 83.9 & 73.2 & 72.9 \\
	GPT-5.5 & None & 74.0 & 82.4 & 70.1 & 71.4 \\
	GPT-5.5 & Low & 90.9 & 91.7 & 89.9 & 88.5 \\
	GPT-5.5 & Medium & 92.6 & 94.9 & 87.2 & 87.2 \\
	GPT-5.5 & High & 93.4 & 95.8 & 89.2 & 89.4 \\
	Gemma 4 (4B) & None & 47.0 & 49.0 & 47.1 & 46.7 \\
	Gemma 4 (4B) & Thinking & 36.1 & 45.8 & 35.1 & 33.4 \\
	Gemma 4 (31B) & None & 66.2 & 73.7 & 67.7 & 64.0 \\
	Qwen3.5 (9B) & None & 35.3 & 41.9 & 29.1 & 31.1 \\
	Qwen3.5 (9B) & Thinking & 57.1 & 53.7 & 52.5 & 54.0 \\
	Qwen3.6 (35B) & None & 60.7 & 59.4 & 56.3 & 61.1 \\
	Qwen3.6 (35B) & Thinking & 65.3 & 64.0 & 56.9 & 55.2 \\
	Llama 3.2 (3B) & None & 11.6 & 13.0 & 13.5 & 13.2 \\
	Llama 3.3 (70B) & None & 36.7 & 44.9 & 35.0 & 34.8 \\
	LLaMarine & None & 35.1 & 43.0 & 32.8 & 33.1 \\
\end{longtable}
\endgroup

The results show a consistent performance gradient across all models and both sections, one that maps directly onto safety roles. Packer/Cargo
Handler scores are highest in virtually every configuration, followed by
Consignor/Freight Forwarder, with Ship Loader and Ship Operator trailing
by a meaningful margin. On Section 1, the gap between Packer/Cargo
Handler and Ship Loader averages roughly 10-12 percentage points across
mid-tier commercial models: for example, Claude Sonnet 4.6 (Max) scores
82.3\% on Packer-relevant content but only 70.8\% on Ship Loader
content, falling below the 75\% pass mark for the two operational roles
despite exceeding it overall. This pattern directly reflects the
subsection difficulty structure: Packer and Consignor curricula are
weighted towards classification and marking questions, where models
perform well, while Ship Loader and Ship Operator curricula carry
heavier weighting on stowage and on-vessel segregation, which are the
consistently weakest areas across all models and sections. The same
gradient appears in Section 2, though the role gap is somewhat narrower.
The safety implication is direct: the roles closest to physical danger, those responsible for loading and operating vessels carrying dangerous cargo, are the roles on which current LLMs perform worst.

\hypertarget{cost}{%
\subsection{5.4 Cost and Latency}\label{cost}}

\hyperlink{tab:4}{Table 4} reports evaluation cost and wall-clock time per 1,000 questions for each model configuration.

\hypertarget{tab:4}{}%
\textbf{Table 4: Evaluation cost and wall-clock time per 1,000 questions, per model configuration.} Costs are only shown for commercial models as open-source models were run
locally and incurred no API cost.

\begingroup\footnotesize
\begin{longtable}[]{@{}lllr@{}}
\toprule\noalign{}
Model & Thinking & Cost / 1{,}000 Q (\$) & Time / 1{,}000 Q \\
\midrule\noalign{}
\endhead
\bottomrule\noalign{}
\endlastfoot
Claude Sonnet 4.6 & None & 0.70 & 42 min \\
Claude Sonnet 4.6 & Low & 0.61 & 40 min \\
Claude Sonnet 4.6 & Medium & 1.69 & 1 h 1 min \\
Claude Sonnet 4.6 & High & 6.43 & 2 h 21 min \\
Claude Sonnet 4.6 & Max & 15.91 & 4 h 57 min \\
Claude Opus 4.7 & None & 1.34 & 54 min \\
Claude Opus 4.7 & Low & 1.22 & 47 min \\
Claude Opus 4.7 & Medium & 1.23 & 45 min \\
Claude Opus 4.7 & High & 1.64 & 50 min \\
Claude Opus 4.7 & Max & 10.09 & 1 h 54 min \\
Gemini 3.1 Flash-Lite & Minimal & 0.04 & 15 min \\
Gemini 3.1 Flash-Lite & Low & 0.24 & 21 min \\
Gemini 3.1 Flash-Lite & Medium & 0.53 & 34 min \\
Gemini 3.1 Flash-Lite & High & 1.25 & 59 min \\
Gemini 3.1 Pro & Low & 4.00 & 1 h 26 min \\
Gemini 3.1 Pro & Medium & 5.58 & 1 h 50 min \\
Gemini 3.1 Pro & High & 8.58 & 2 h 33 min \\
GPT-5.4 Mini & None & 0.14 & 23 min \\
GPT-5.4 Mini & Low & 1.94 & 1 h 0 min \\
GPT-5.4 Mini & Medium & 9.20 & 3 h 21 min \\
GPT-5.4 Mini & High & 25.97 & 5 h 44 min \\
GPT-5.5 & None & 0.92 & 36 min \\
GPT-5.5 & Low & 11.41 & 2 h 11 min \\
GPT-5.5 & Medium & 29.93 & 4 h 10 min \\
GPT-5.5 & High & 69.75 & 8 h 48 min \\
Gemma 4 (4B) & None & - & 7 min \\
Gemma 4 (4B) & Thinking & - & 38 min \\
Gemma 4 (31B) & None & - & 5 min \\
Llama 3.2 (3B) & None & - & 10 min \\
Llama 3.3 (70B) & None & - & 21 min \\
Qwen3.5 (9B) & None & - & 4 min \\
Qwen3.5 (9B) & Thinking & - & 12 h 14 min \\
Qwen3.6 (35B) & None & - & 4 min \\
Qwen3.6 (35B) & Thinking & - & 3 h 6 min \\
LLaMarine & None & - & 15 min \\

\end{longtable}
\endgroup

GPT models account for the majority of commercial spend: GPT-5.5 alone accounts for more than half of the total across its four thinking configurations, with its High configuration exceeding one-third of the entire commercial budget. The rate at which thinking level increases cost varies substantially between providers. For GPT-5.5, the High configuration cost approximately 75 times more than the None configuration; for GPT-5.4 Mini the ratio is approximately 190 to one. Within Claude, Sonnet at Max cost 23 times more than Sonnet at None. Gemini models are substantially cheaper: the entire Gemini evaluation across all seven Flash-Lite and Pro configurations cost less than a single GPT-5.5 High run. Increasing thinking level consistently increases cost across all providers.

Latency broadly mirrors the cost rankings, and the figures here are likewise reported per 1,000 questions. Gemini 3.1 Flash-Lite is the fastest commercial family: the Minimal configuration processed 1,000 questions in approximately 15 minutes and the High configuration in under an hour. Gemini 3.1 Pro at High thinking required approximately 2.6 hours per 1,000 questions. GPT models are consistently the slowest: GPT-5.5 at High thinking took approximately 8.8 hours per 1,000 questions, around 3.5 times longer than Gemini 3.1 Pro High despite lower accuracy on three of four sections. Within every commercial family, increasing thinking level increases latency monotonically. For locally-hosted open-source models, no-thinking inference is fast (roughly 4-21 minutes per 1,000 questions depending on model size), but thinking configurations vary: Qwen3.5 (9B) required approximately 12.2 hours per 1,000 questions, likely reflecting very long generated reasoning chains rather than model size.

\hypertarget{web-search}{%
\subsection{5.5 Web Search}\label{web-search}}

Most consumer AI assistants, including ChatGPT and Gemini, have web search enabled by default, so practitioners using these tools typically receive responses augmented with live search results rather than training data alone.
How much search improves DG task performance is therefore a practical question for anyone using these tools.
When web search is enabled, the model retrieves pages from the internet at inference time and incorporates them into its context before generating a response.
This can compensate for training data gaps but increases input length and cost.

Web search has dramatically different effects depending on task type. On
Section 3 (DGL lookup), the average improvement across the three models
is +25.0 percentage points, consistent with
the DGL being accessible via web retrieval. GPT-5.4 Mini shows the largest
benefit (+49.6 pp), rising from 32.2\% to 81.9\%; approaching the
no-search performance of large commercial models. On Section 1 (MCQ),
the average change is effectively zero (-0.3 pp), suggesting that answer
retrieval adds no benefit when questions require reasoning over
multiple-choice options that are already provided.

\hyperlink{tab:5}{Table 5} presents the impact of enabling web search, tested on a subset
of three small commercial models (None/Minimal thinking) on Sections 1
and 3.

\hypertarget{tab:5}{}%
\textbf{Table 5: Impact of web search on selected models.}

\begingroup\footnotesize
\begin{longtable}[]{@{}llrrrr@{}}
\toprule\noalign{}
Section & Model & No Search (\%) & Search (\%) & Change (pp) & Cost Multiplier \\
\midrule\noalign{}
\endhead
\bottomrule\noalign{}
\endlastfoot
1 & Claude Sonnet 4.6 (None) & 57.9 & 51.0 & -6.9 & 194× \\
1 & Gemini 3.1 Flash-Lite (Minimal) & 65.9 & 72.2 & +6.3 & 7× \\
1 & GPT-5.4 Mini (None) & 53.6 & 53.2 & -0.4 & 55× \\
3 & Claude Sonnet 4.6 (None) & 63.7 & 69.3 & +5.6 & 538× \\
3 & Gemini 3.1 Flash-Lite (Minimal) & 67.9 & 87.8 & +19.9 & \textasciitilde1× \\
3 & GPT-5.4 Mini (None) & 32.2 & 81.9 & +49.6 & 45× \\
\end{longtable}
\endgroup

Cost-efficiency effects are sharply model-dependent. Gemini Flash-Lite
achieved a +19.9 pp improvement on Section 3 at negligible additional
cost (cost multiplier approximately 1×, as searches appear to have been
served from cache or at very low cost). In contrast, enabling web search
for Claude Sonnet 4.6 on Section 3 increased cost by a factor of 538×,
with only a 5.6 pp improvement. GPT-5.4 Mini achieved the largest
accuracy gain at a cost multiplier of 45×. These results suggest that
for practitioners accessing DG lookup assistance via consumer chatbots
with search enabled, small models can achieve performance approaching
that of large no-search models on retrieval-oriented tasks. The costs
stated here are in addition to those in \hyperref[cost]{Section 5.4}.

\hypertarget{discussion}{%
\section{6. Discussion}\label{discussion}}

The discussion that follows should be read against the study's central
finding: no model is yet reliable enough to act without oversight, and even
the strongest performer is weakest where errors are most dangerous.

\hypertarget{model-recommendations-for-practitioners}{%
\subsection{6.1 Model Performance for
Practitioners}\label{model-recommendations-for-practitioners}}

Gemini 3.1 Pro achieves the highest scores across all four benchmark sections.
It is the only model to substantially
exceed the human practitioner baseline on Section 1 (95.8\% vs 83.8\%)
and it leads by wide margins on Sections 2 and 3.
It has meaningful weaknesses, however: stowage codes, segregation, special provisions, and regulatory recall (Section 4) remain challenging even for the best-performing model, and practitioners should exercise particular caution when using any LLM for queries in these areas.

GPT-5.5 (at Low thinking and above) is the only other model family to
exceed the human baseline on Section 1, though it trails Gemini 3.1 Pro on every metric while costing
substantially more. Within the Gemini 3.1 Pro configurations, minimal
thinking (the Low setting) gives strong cost-performance: Section 1 at 94.4\%
for \$4.00 per 1,000 questions, compared to 95.8\% at \$8.58 for High thinking. For ad-hoc DGL
lookups, Gemini Flash-Lite or GPT Mini with web search enabled score substantially better on retrieval-oriented tasks at lower cost than larger no-search models.

\hypertarget{thinking-mode-effects}{%
\subsection{6.2 Thinking Mode Effects}\label{thinking-mode-effects}}

The relationship between increased thinking and performance is
inconsistent across models and sections. Most models show moderate
improvements with greater thinking, but several show ``overthinking''
penalties at high thinking levels:

\begin{itemize}
\tightlist
\item
  GPT-5.4 Mini deteriorates from High thinking on Sections 3 and 4
  relative to Medium (Section 3: 47.5\% → 46.4\%; Section 4: 15.6\% →
  12.6\%).
\item
  Claude Sonnet 4.6 shows declining performance on Sections 2 and 4 as
  thinking increases beyond Low (Section 2: 60.4\% → 57.8\% at Low, then
  partial recovery).
\item
  Claude Sonnet 4.6 Max achieves the best Section 1 result for that
  model family, but at the lowest Section 4 performance of the family
  (7.5\%).
\end{itemize}

Gemini 3.1 Pro is the most consistent beneficiary of thinking: all three
thinking levels produce similar and very high Section 1 scores, with a
small monotonic improvement on Section 4 (34.1\% → 35.0\% → 36.8\%).

\hypertarget{google-models-and-training-data}{%
\subsection{6.3 Google Models and Training
Data}\label{google-models-and-training-data}}

The dominance of Google models (both Gemini 3.1 Pro and, to a lesser
extent, Gemini Flash-Lite and Gemma 4-31B) across all sections suggests
superior coverage of dangerous goods regulatory content in Google's
pre-training corpora. The full IMDG Code is distributed through paid
official print and electronic editions rather than as an openly downloadable
digital corpus, making systematic inclusion in ordinary web crawls less likely.
Google's advantage may therefore reflect broader coverage of
derivative sources, industry guidance documents, IMO circulars, company
safety procedures, training materials, and commentary that reference or
paraphrase IMDG provisions, or possibly licensed access to regulatory
content. Training data composition is undisclosed across all providers,
so the source of Google's advantage cannot be determined.
The pattern of weak areas is consistent with the hypothesis that training sources do not systematically enumerate operational fields such as
stowage codes, segregation groups, and transport operation procedures. These remain hard for all models including Gemini, suggesting a ceiling imposed by source availability rather than model capability.

\hypertarget{safety-guardrails-and-refusals}{%
\subsection{6.4 Safety Guardrails and
Refusals}\label{safety-guardrails-and-refusals}}

Some questions in DGEval were refused by models on safety grounds,
resulting in a score of zero. Claude Sonnet 4.6 and Claude Opus 4.7
refused to answer four questions across all thinking levels:
specifically questions about pathogens, inhalation toxicity, and
dibromomethane. Gemini 3.1 Flash-Lite and GPT-5.5 also refused to answer the
dibromomethane question; GPT-5.5 additionally refused to answer a question about
nitroglycerin. These refusals are a systematic bias that
disproportionately affects certain safety-critical DG topics. An LLM
that cannot distinguish between an adversarial query and a legitimate
regulatory compliance question has limited utility in a DG context:
the inability to answer questions about inhalation toxicity or explosive
substances is a capability gap that makes the model
unreliable for DG practitioners.

\hypertarget{limitations}{%
\subsection{6.5 Limitations}\label{limitations}}

\textbf{Knowledge cutoff sensitivity.} IMDG A42-24 came into force on 1
January 2025; models trained before this date may lack specific A42-24
content. However, changes between consecutive amendments
are small; we estimate A42-24 changed less than 0.5\% of the content relative to A41-22.
By design, questions tied explicitly to a given amendment are kept in a
dedicated IMDG refresher course rather than the generic exam bank from which
DGEval is drawn. The benchmark
therefore contains few explicitly A42-24-specific questions, so the residual
effect of cutoff differences on our results is limited, though we do not
quantify it precisely.

\textbf{Benchmark contamination.} The NCB Hazcheck e-learning content is not
publicly distributed, and as noted above, the IMDG Code is not
available as an open digital corpus. Both factors substantially
reduce the contamination risk relative to benchmarks constructed from freely
accessible sources, and the controlled-release evaluation model described in
\hyperref[appendix-e-benchmark-access]{Appendix E} is intended to keep this risk low for future models as well. That
said, contamination cannot be ruled out entirely: NCB
Hazcheck publishes some materials publicly (including amendment summaries and
guidance documents), secondary sources that paraphrase or derive from the
IMDG Code are widely available, and it is unknown whether any model provider
has licensed regulatory content. The open-ended Section 2 remains somewhat
more robust to any residual contamination than MCQ Section 1, since producing
a correct free-text answer requires generative reasoning rather than
selecting a memorised option.

\textbf{IMDG scope only.} This benchmark covers only the IMDG Code. The
US CFR Title 49 (domestic US) and dangerous goods regulations for air
(IATA), road (ADR) and rail (RID) transport are all out of scope. These present
distinct regulatory texts and may warrant separate benchmarks.

\textbf{Coverage gaps.} Stowage and segregation subsections show the weakest
performance across all sections and carry the highest danger-criticality weights per expert review.
The combination of poor model performance and high operational risk makes these areas the most
important to investigate further in future work.

Whilst this benchmark focuses on textual knowledge, a significant part of
DG regulation is concerned with ensuring goods are marked, packed and
stowed securely. This can only be done through visual inspection.
A multimodal extension of DGEval, drawing on images of placards, labels,
and stowage arrangements, would therefore be a natural direction for future work.

\textbf{Practitioner-query realism.} The benchmark tests regulatory
knowledge one well-formed question at a time. Real compliance is not made
this way: a single consignment passes through consignors, packers,
forwarders, terminal operators, and ship operators, whose decisions on
classification, packing, segregation, and stowage depend on one another.
Real queries are also messier, often ambiguous, tied to a specific
commodity, or dependent on information the Code does not contain, such as
how a container was packed, what a terminal can accept, or how a vessel is
stowed. A benchmark of isolated questions understates this difficulty, so a
high score should not be read as a measure of operational decision quality.
A useful next step would be scenario-based evaluation that scores whole
multi-step workflows rather than single questions.

\hypertarget{safe-deployment}{%
\subsection{6.6 Safe Deployment and Continuous Assurance}\label{safe-deployment}}

The results position LLMs as safety-support tools rather than safety authorities.
Top-performing models can help practitioners access and interpret complex regulatory material, and in some areas approach or exceed trained human baselines.
No tested model achieves 100\% accuracy across any section, however, and the nature of errors matters as much as their frequency.

Performance is highly uneven across sub-domains.
Stowage, segregation, and regulatory recall show scores substantially below the section means, and these carry the highest danger-criticality weights per expert review.
A composite score is therefore insufficient as a deployment criterion: a model achieving 80\% overall may still be unreliable in the sub-domains where errors carry the most serious consequences.
Deployment decisions should be informed by task-specific performance, not aggregate scores alone.

LLM capabilities also evolve rapidly, and performance varies across model versions even within the same provider family, as the spread across thinking configurations in this study illustrates.
DGEval is not a one-time characterisation of current capability; it is a safety assurance instrument that must be applied continuously.
When models are updated, retired, or replaced, re-evaluation against DGEval gives practitioners and their organisations current, evidence-based judgements about model reliability on DG compliance tasks.

Safe deployment requires embedding LLMs within a governed workflow where trained practitioners retain approval authority.
Using an LLM for first-pass lookup and orientation, then verifying against the IMDG Code or a certified compliance system, gives practitioners the efficiency of AI assistance without delegating safety decisions to a model that remains fallible in critical areas.
Human oversight is not a transitional measure pending better models; it is a structural requirement until models consistently score 100\% across all operationally sensitive sub-domains. The governance and liability questions raised by delegating maritime safety decisions to autonomous and AI-driven systems are themselves an active area of study~\citep{chen2023intelligentships}, and they reinforce the case for retaining a human decision-maker in the DG compliance loop.

\hypertarget{conclusion}{%
\section{7. Conclusion}\label{conclusion}}

This paper presented DGEval, the first LLM benchmark grounded
specifically in IMDG A42-24 dangerous goods shipping
regulation. Across 1,678 questions in four sections, evaluated on 13
models and 35 configurations, Gemini 3.1 Pro consistently dominates all
other tested models. On Section 1 (multiple-choice), it substantially
exceeds the human practitioner baseline of 83.8\% at all tested thinking
levels, with GPT-5.5 being the only other model to do so. Section 4
(regulatory recall: identifying the specific IMDG section containing an
answer) is uniformly difficult for all models, with a mean score of
14.2\%. Stowage and segregation are consistent weak points across all
models and sections.

Web search substantially improves DGL lookup accuracy (+25.0 pp on
average for Section 3), with cost-efficiency varying dramatically by
model. The maritime-specific LLaMarine model performs no better than
base Llama 3.3 (70B) on DG questions, confirming the need for
DG-specific training or evaluation data.

Among tested models, Gemini 3.1 Pro achieves the best results across all benchmark sections, though caution is warranted in stowage, segregation, and regulatory-text recall regardless of model choice.
For budget-constrained or high-volume lookup tasks, Gemini Flash-Lite with web search enabled scores substantially better than its no-search baseline at low additional cost.
None of the models evaluated here are validated or certified for operational DG compliance decisions, and nothing in this paper should be read as a recommendation to use one for that purpose.

More broadly, the results expose a gap between dangerous goods knowledge and dangerous goods decision-making, and support treating LLMs as decision-support tools rather than safety authorities.
Even the best-performing model is not uniformly reliable, and composite scores mask sub-domain failures in the areas of highest operational risk.
Deployment should be within governed workflows where trained practitioners retain approval authority and verify outputs against the IMDG Code or a certified compliance system.
Aggregate scores should not be the sole basis for deployment decisions; task-specific performance in safety-critical sub-domains is the relevant criterion.
DGEval is not a point-in-time verdict on current models: it is a safety assurance instrument to be applied on an ongoing basis as capabilities change.

Future work will investigate stowage and segregation further, given that
these subsections show the weakest model performance and carry the highest
danger-criticality weights, and explore multimodal evaluation to address
the visual inspection aspects of DG compliance that text-only tasks cannot capture.
A further direction is scenario-based evaluation that scores whole
multi-step compliance workflows rather than isolated questions, bringing
the benchmark closer to how decisions are made in practice.

\newpage
\hypertarget{appendices}{%
\section{Appendices}\label{appendices}}

\hypertarget{appendix-a-evaluation-metrics-detail}{%
\subsection{Appendix A: Evaluation Metrics
Detail}\label{appendix-a-evaluation-metrics-detail}}

\textbf{Partial credit.} For questions with \(n\) correct answers where
the model returns \(m\) answers, of which \(k\) are correct, the score
is:

\[\text{score} = \frac{k}{\textrm{max}(n,m)}\]

This penalises over-prediction symmetrically with under-prediction.

\textbf{Subsection weighting (Sections 1, 2, 4).} Let \(q_i\) be the
score for question \(i\) in subsection \(s\), and \(w_s\) the normalised
weight for subsection \(s\). The section score is:

\[\text{Section score} = \sum_s w_s \cdot \bar{q}_s\]

where \(\bar{q}_s\) is the mean score across all questions in subsection
\(s\). Subsection weights can be found in \hyperref[appendix-b-e-learning-subsection-details]{Appendix B}.

\textbf{DGL column weighting (Section 3).} The same formula applies,
with DGL columns in place of subsections.
Subsection weights can be found in \hyperref[appendix-c-dgl-column-descriptions-and-weights]{Appendix C}.

\textbf{Section-level composite score.}
For the final benchmark score, each of the four sections was given equal weight, and the section results averaged.

\hypertarget{appendix-b-e-learning-subsection-details}{%
\subsection{Appendix B: E-learning Subsection
Details}\label{appendix-b-e-learning-subsection-details}}

\hypertarget{tab:a1}{}%
\textbf{Table A1: E-learning subsections with weights and practitioner
relevance.} C/FF: Consignor/Freight Forwarder; P/CH: Packer/Cargo Handler; SL: Ship Loader; SO: Ship Operator.

\begingroup\footnotesize
\begin{longtable}[c]{llrrl}
\toprule\noalign{}
Code & Title & Score & Weight & Relevant Roles \\
\midrule\noalign{}
\endhead
\bottomrule\noalign{}
\endlastfoot
E51 & General Provisions & 4 & 0.027 & All \\
E52 & Classification - Substances \& PSN & 6 & 0.041 & P/CH; SL \\
E53 & Packing Provisions & 6 & 0.041 & P/CH; SL \\
E54 & Consignment - Marks \& Documents & 6 & 0.041 & Standard \\
E55 & Transport Operations - Stowage \& Segregation & 6 & 0.041 & Standard \\
E55b & Transport Operations - Container Safety \& Emergency & 6 & 0.041 & SL \\
E56 & Documentation & 6 & 0.041 & Standard \\
E60 & Classification - Procedures \& Code & 10 & 0.068 & C/FF; SO \\
E61 & Additional Classification Considerations & 7 & 0.048 & C/FF; SO \\
E62 & Packing Provisions for Packagings & 8 & 0.054 & C/FF \\
E63 & Packing Provisions for Tanks and Bulk Containers & 7 & 0.048 & C/FF \\
E64 & Consignment - Detailed Provisions & 10 & 0.068 & C/FF; P/CH \\
E65 & Stowage of Dangerous Goods & 9 & 0.061 & SL; SO \\
E66 & Segregating Dangerous Goods within a TEU & 10 & 0.068 & SL; SO \\
E67 & Segregating Dangerous Goods on Board Vessels & 9 & 0.061 & SL; SO \\
E68 & Limited and Excepted Quantities Provisions & 8 & 0.054 & C/FF; SO \\
E69 & Transport Operations - Containers & 7 & 0.048 & Advanced only \\
E69a & Transport Operations - Containers (Basics) & 7 & 0.048 & SO \\
E70 & Transport Operations - Dangerous Goods in Port Areas & 7 & 0.048 & SO \\
E71 & Transport Operations - Emergency Response & 8 & 0.054 & SO \\
\end{longtable}
\endgroup

Table A1 lists the 20 e-learning subsections evaluated in Sections 1, 2,
and 4, with expert-assigned scores and the normalised weights used in
benchmark scoring. The expert score reflects the danger-criticality of
each subsection as assessed by two in-house NCB Hazcheck DG experts via
consensus. Weights are proportional to scores and normalised to sum to
1. Also shown are the relevances of the subsections to the practitioner
courses.

Note that not all of the subsections are relevant to
practitioner-specific courses. Subsections E54, E55, E56 and E69 are
only used in non-specific (Standard and Advanced) courses.

\hypertarget{appendix-c-dgl-column-descriptions-and-weights}{%
\subsection{Appendix C: DGL Column Descriptions and
Weights}\label{appendix-c-dgl-column-descriptions-and-weights}}

Table A2 lists the 16 DGL properties evaluated in Section 3, with
Hazcheck's descriptions and expert-assigned importance weights. UN Number and PSN
are each weighted at 25\% irrespective of expert scoring; the remaining
14 columns share the remaining 50\%, weighted proportionally to expert
scores.

\hypertarget{tab:a2}{}%
\textbf{Table A2: DGL fields, descriptions, and weights.}

\begingroup\footnotesize
\begin{longtable}[c]{lclrr}
\toprule\noalign{}
Field & DGL Column & Description & Expert Score & Weight \\
\midrule\noalign{}
\endhead
\bottomrule\noalign{}
\endlastfoot
UN Number & 1 & Four-digit UN identifier assigned to a dangerous good & - & 0.250 \\
PSN & 2 & Proper Shipping Name - official transport name & - & 0.250 \\
Primary Hazard Class & 3 & Hazard class, division, and compatibility group & 10 & 0.128 \\
Subsidiary Classes & 4 & Class numbers of any subsidiary hazards & 5 & 0.064 \\
Packing Group & 5 & I (most dangerous) to III (least dangerous) & 5 & 0.064 \\
Emergency Schedule & 15 & Emergency response codes for fire or spillage & 3 & 0.038 \\
Stowage Category & 16a & Vessel stowage category code & 2 & 0.026 \\
Packing Instructions & 8 & Codes for permitted packaging types & 2 & 0.026 \\
Stowage Code & 16a & Stowage location and special handling codes & 2 & 0.026 \\
Segregation Code & 16b & Codes for classes/substances that must be kept apart & 2 & 0.026 \\
Special Provisions & 6 & Additional packaging requirement codes & 2 & 0.026 \\
Limited Quantity & 7a & Maximum quantity per package for partial exemption & 2 & 0.026 \\
Tank Instructions & 13 & T-codes for portable tank transport & 1 & 0.013 \\
Marine Pollutant & 4 & Whether the material is harmful to aquatic life & 1 & 0.013 \\
Segregation Group & 16b & Segregation group classification & 1 & 0.013 \\
Excepted Quantity & 7b & Exemption code for very small quantity shipments & 1 & 0.013 \\
\end{longtable}
\endgroup

\hypertarget{appendix-d-llm-judge-validation}{%
\subsection{Appendix D: LLM Judge
Validation}\label{appendix-d-llm-judge-validation}}

Section 2 responses were evaluated by \texttt{gemini-2.5-flash}
(non-thinking) as an automated LLM judge. To validate the judge's reliability, we conducted a meta-evaluation following \citep{electronics15030659}, where human experts evaluated the quality and alignment of the judge's assessments rather than re-annotating the underlying tasks.

We randomly sampled 30 responses from each of three models, Claude Opus 4.7 (Low Thinking), Gemini 3.1 Pro (Low Thinking), and GPT-5.5 (Low Thinking), for a total of 90 (question, response) pairs. For each pair, the judge assigned a score from 1 to 5 and provided a textual justification across five quality dimensions: Factual Accuracy, Completeness, Relevance and Conciseness, Clarity and Readability, and Overall Coherence. This process produced 450 evaluation records (90 pairs × 5 dimensions).

From these 450 records, we randomly sampled 100 for human review. Two in-house DG experts independently evaluated the same set of records and provided:

\begin{itemize}
\tightlist
\item \textbf{Q1 (Score Validity):} A 1-5 rating indicating the extent to which the expert agreed with the judge's assigned score.
\item \textbf{Q2 (Explanation Quality):} A categorical assessment of the judge's justification as \emph{precise}, \emph{plausible}, or \emph{incorrect}.
\end{itemize}

We report three metrics to quantify alignment between the judge and human experts:

\begin{itemize}
\tightlist
\item \textbf{Judgment Acceptance Rate (JAR):} The percentage of cases where Q1 $\geq$ 4, indicating broad expert agreement with the judge's assessment.
\item \textbf{Explanatory Utility Index (EUI):} The percentage of cases where Q2 is labeled \emph{precise}.
\item \textbf{Cohen's Kappa:} Inter-expert agreement on Q1 and Q2, measuring the reliability of the human evaluation process.
\end{itemize}

\begin{table}[!htbp]
\centering
\hypertarget{tab:a3}{}%
\textbf{Table A3: Overall judge performance.}\\[4pt]
\begin{tabular}{lcccc}
\toprule
Expert & JAR (\%) & 95\,\% CI & EUI (\%) & 95\,\% CI \\
\midrule
Expert~1 & 74.00 & [64.63,\,81.60] & 73.00 & [63.57,\,80.73] \\
Expert~2 & 71.00 & [61.46,\,78.99] & 78.00 & [68.93,\,85.00] \\
\bottomrule
\end{tabular}
\end{table}

\begin{table}[!htbp]
\centering
\hypertarget{tab:a4}{}%
\textbf{Table A4: Agreement between the two experts.}\\[4pt]
\setlength{\tabcolsep}{4pt}
\small
\begin{tabular}{@{}llcccccc@{}}
\toprule
Variable & Kappa Type & $\kappa$ & Interpretation
         & $P_a$ (\%) & $P_e$ (\%) & $n$ \\
\midrule
Q1 & Quadratic-weighted & 0.7932 & Substantial
                          & 96.19 & 81.56 & 100 \\
Q2 & Nominal            & 0.1228 & Slight
                          & 66.00 & 61.24 & 100 \\
\bottomrule
\end{tabular}
\end{table}

\hyperlink{tab:a3}{Table A3} and \hyperlink{tab:a4}{Table A4} report the judge performance and inter-expert agreement. The judge achieved high alignment with human experts, with JAR of 74.00\% (95\% CI [64.63, 81.60]) and 71.00\% (95\% CI [61.46, 78.99]) for Experts~1 and~2, respectively. The judge's explanations were rated as \emph{precise} in 73.00\% and 78.00\% of cases (EUI), with overlapping confidence intervals indicating consistent evaluations.

Inter-expert agreement was substantially higher for score validity than for explanation quality. For score validity, experts showed strong agreement ($\kappa = 0.7932$, $P_a = 96.19\%$), whereas agreement on explanation quality was much lower ($\kappa = 0.1228$, $P_a = 66.00\%$), despite non-negligible observed agreement driven by high chance agreement ($P_e = 61.24\%$) and a skewed label distribution. Overall, the strong JAR and EUI results, plus substantial agreement on Q1, support the reliability of the LLM judge.

\hypertarget{appendix-e-benchmark-access}{%
\subsection{Appendix E: Benchmark
Access}\label{appendix-e-benchmark-access}}

DGEval uses a controlled-release evaluation model. The benchmark questions
and answer key are withheld to prevent future models from being trained
directly on the test set, which would undermine the benchmark's long-term
value as an independent evaluation tool. This follows the practice of
established benchmarks: nuScenes~\citep{caesar2020nuscenes}, for example,
allows participants to submit predictions to an evaluation server and
receive official scores without direct access to the hidden test labels.

We plan to provide a submission-based evaluation server following this
model: access will require signing a licence agreement, and submitting
model answers will return benchmark scores without exposing the underlying
questions or answers. Details are available on request from the authors.

Access to DGEval does not grant any right to reproduce, redistribute or
provide access to IMO publications. Users remain responsible for obtaining
any licences required for their own access to or use of the IMDG Code.

\newpage
\bibliographystyle{plainnat}
\bibliography{references}

\end{document}